\documentclass[a4paper,twoside]{article}

\usepackage{epsfig}
\usepackage{subcaption}
\usepackage{calc}
\usepackage{amssymb}
\usepackage{amstext}
\usepackage{amsmath}
\usepackage{amsthm}
\usepackage{multicol}
\usepackage{pslatex}
\usepackage{apalike}
\usepackage{algorithm2e}
\usepackage[bottom]{footmisc}

\usepackage[utf8]{inputenc} % allow utf-8 input
\usepackage[T1]{fontenc}    % use 8-bit T1 fonts
\usepackage[hidelinks]{hyperref}       % hyperlinks
\usepackage{url}            % simple URL typesetting
\usepackage{booktabs}       % professional-quality tables
\usepackage{amsfonts}       % blackboard math symbols
\usepackage{nicefrac}       % compact symbols for 1/2, etc.
\usepackage{microtype}      % microtypography
\usepackage{xcolor}         % colors
\usepackage{subcaption}
\usepackage{graphicx}

\usepackage{SCITEPRESS}     % Please add other packages that you may need BEFORE the SCITEPRESS.sty package.

\begin{document}

\title{Explainability-Driven Leaf Disease Classification Using Adversarial Training and Knowledge Distillation}

\author{\authorname{Sebastian-Vasile Echim\sup{1,}\sup{*}, Iulian-Marius Tăiatu\sup{1,}\sup{*}, Dumitru-Clementin Cercel\sup{1,}\thanks{Corresponding author}, and Florin Pop\sup{1,2,3}}
\affiliation{\sup{1} Computer Science and Engineering Department, National University of Science and Technology POLITEHNICA Bucharest, Romania}
\affiliation{\sup{2} National Institute for Research and Development in Informatics - ICI Bucharest, Romania}
\affiliation{\sup{3} Academy of Romanian Scientists, Bucharest, Romania}
\email{dumitru.cercel@upb.ro}
}


\abstract{This work focuses on plant leaf disease classification and explores three crucial aspects: adversarial training, model explainability, and model compression. The models' robustness against adversarial attacks is enhanced through adversarial training, ensuring accurate classification even in the presence of threats. Leveraging explainability techniques, we gain insights into the model's decision-making process, improving trust and transparency. Additionally, we explore model compression techniques to optimize computational efficiency while maintaining classification performance. Through our experiments, we determine that on a benchmark dataset, the robustness can be the price of the classification accuracy with performance reductions of 3\%-20\% for regular tests and gains of 50\%-70\% for adversarial attack tests. We also demonstrate that a student model can be 15-25 times more computationally efficient for a slight performance reduction, distilling the knowledge of more complex models.}

\onecolumn \maketitle \normalsize \setcounter{footnote}{0} 

\def\thefootnote{*}\footnotetext{Equal contributions.}\def\thefootnote{\arabic{footnote}}

\section{\uppercase{Introduction}}
% For many decades, artificial intelligence has been a field heavily focused on theory with applications of real-world impact. 

Artificial intelligence (AI) has provided the theoretical knowledge to develop real-world applications for decades. The amalgamation of enhanced computational capacities, refined learning algorithms, and increased data accessibility has propelled significant advances in machine learning, resulting in the use in many domains \cite{linardatos2021xai}. Furthermore, the advancement of deep learning techniques, demonstrated by their ongoing refinement, has reinforced their pivotal role in the AI landscape \cite{he2016residual,khan2022deep}.

Nevertheless, although deep neural networks (DNNs) can attain best performance for various tasks in domains such as legal \cite{smuadu2022legal},  energy \cite{nuastuasescu2022conditional}, and medicine \cite{lungu2023skindistilvit}, it has been demonstrated that these networks are prone to classification errors for data with imperceptible noise \cite{szegedy2014intriguing,goodfellow2015fgsm}. Thus, the network sensitivity to perturbations is explored using adversarial algorithms \cite{goodfellow2015fgsm,kurakin2016bim}. 
They are used in network attacks, robustness improvements, or data augmentations to achieve better performance with adversarial training \cite{miyato2016adversarial}.

The prediction accuracy is frequently attained by elevating DNN architecture complexity \cite{linardatos2021xai,meske2022xai}. However, boosting a model's predictive power reduces its ability to explain its inner workings and mechanisms. Consequently, the rationale behind their decisions becomes quite challenging to understand, and, therefore, their predictions take time to interpret. Considerable emphasis has been placed on elucidating the explainability of deep learning, particularly in image analysis, specifically focusing on the concept of saliency maps \cite{ramirez2013saliency,simonyan2014deepic}.

We conduct a series of experiments on a publicly available dataset \cite{pandian2019data} to better understand DNNs and their performance in the context of adversarial attacks and defenses \cite{szegedy2014intriguing}, explainability \cite{simonyan2014deepic}, and knowledge distillation \cite{bucila}. We bring in this work the following main contributions:
\begin{itemize}
   \item We demonstrate the performance impact on DNNs when extra training data is generated with adversarial algorithms for achieving better robustness.
   \item We indicate the effectiveness of adversarial attacks on models not trained with adversarial data and the defense capability when adversarial training is also performed. 
   \item We show that different shapes and orientations of a plant leaf to be classified can determine different results when using adversarial attacks. 
   \item We offer a closer look into the attentional patterns, demonstrating that models can have significant focus shifts away from the main classification points of interest when attacked.
   \item We find a good trade-off solution between accuracy and efficiency using knowledge distillation, obtaining only 13-15\% classification accuracy reduction for models 2-14 times smaller and 15-25 times more efficient.
\end{itemize}

The paper is structured in six sections, starting with the introduction and the related work. The third section describes the dataset characteristics. The fourth section covers the methods used in our experiments, indicating the network architectures employed, the adversarial algorithms applied, the performance metrics used to analyze the results, and the main research questions on which we based our experiments. Finally, the work covers the result analysis in Section 5 and closes with a conclusion section.

\section{\uppercase{Related Work}}

Geetharamani and Pandian \cite{pandian2019data} proposed a leaf disease identification model based on DNNs. The authors train their model using a dataset with 39 different classes of plant leaves. This predictive model achieved impressive results, boasting a 96.46\% accuracy in classification. This performance surpasses classical machine learning baselines. They anticipate even better results with a more complex model incorporating adversarial training. In this experimental area, You et al. \cite{haotian2023adversarial} evaluated a set of adversarial algorithms to attack neural networks such as ResNet50 \cite{he2016residual} and VGG16. Moreover, using the adversarial examples as data augmentation, the authors achieved performance improvements.

Recently, knowledge distillation has become a significant strategy in plant leaf disease recognition, facilitating the creation of compact and practical models. For example, Ghofrani et al. \cite{ghofrani2022knowledge} introduced a client-server system based on the knowledge distillation technique to enhance the client's accuracy. In their approach, a robust deep convolutional neural network (CNN) \cite{kim2014convolutional} architecture was responsible for classifying diseases on the server side. In contrast, on the client side, a lightweight deep CNN architecture facilitated classification on mobile devices.

Musa et al. \cite{musa2022knowledgedistill} explored efficient mobile solutions for AI in agriculture, implementing a low-power DNN for plant disease diagnosis using knowledge distillation. Their teacher and student models are based on the CNN architecture, featuring around 52M parameters for the teacher model and around 1.2M for the student. The system was deployed on a Raspberry Pi device, achieving a 90\% power reduction, with a maximum of 6.22W, and a top performance of plant disease detection of 99.4\% classification accuracy. 
Likewise, Huang et al. \cite{huangkd} tackled the issue of creating small plant disease classification systems based on knowledge distillation, applicable across various crops. Their solution, called multistage knowledge distillation, upholds performance and reduces the model size. Furthermore, the authors underscored the possibility of extending this technique to diverse computer vision tasks, such as image classification and segmentation, enabling the development of automated plant disease detection models with broader utility in precision agriculture.

\section{\uppercase{Dataset}}
In this paper, we employ a leaf disease dataset \cite{pandian2019data} consisting of images of the leaves of different plants such as apple, cherry, corn, grape, pepper bell, and potato (38 different classes). Moreover, the dataset introduces an extra "false" class, covering several landmark photos without leaves.

\begin{figure}[!htb]
  \centering
  \includegraphics[width=0.48\textwidth]{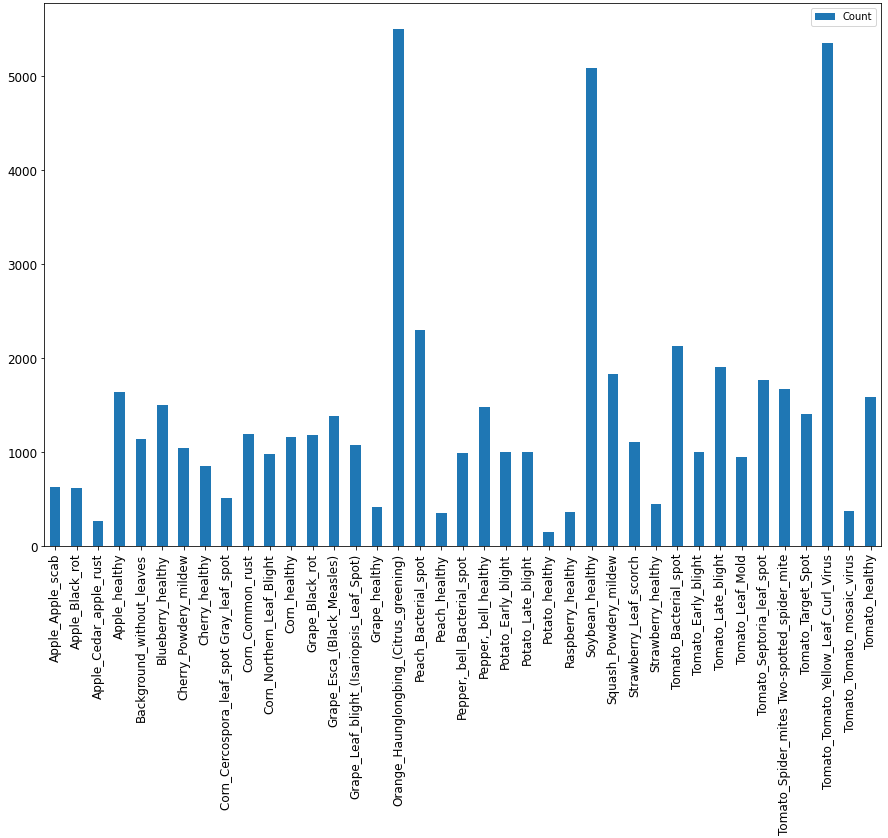}
  \caption{Class distribution of the leaf disease dataset.}
  \label{fig:1-data}
\end{figure}

The dataset contains 55,448 samples with RGB, 256x256 pixels each. As shown in Figure~\ref{fig:1-data}, the dataset is highly imbalanced, with sizes from 100 to over 5,500 per class, thus imposing a classification challenge due to various leaf disease types. The classes with the most significant training footprint are Orange with Huanglongbing, Healthy soybean, and Tomato with yellow leaf curl virus. We expect a better classification accuracy for these categories. In contrast, classes such as Healthy potato, Healthy raspberry, and Apple with cedar apple rust will most likely be unfavoured classification-wise.

\section{\uppercase{Methodology}}

\subsection{Model}

The architecture consists of a regular CNN model as a feature extractor built based on different filters with (3, 32), (32, 64), (64, 128), (128, 256) features with ReLU activations and max-pooling, as well as a classifier represented by two fully-connected layers with 128 inputs to 39 outputs, that is, the number of classes for our dataset. The optimizer is Adam \cite{adam2015diederik}, whose learning rate is 5e-4 with linear decay, while the loss function is the cross-entropy.

\subsection{Adversarial Training and Attack}

Regarding adversarial attacks and training, several algorithms were introduced in our experiments (see Table~\ref{tab:1-adversarial}). We use these gradient-based algorithms to test the performance of the CNN model under several adversarial configurations. The experiments performed in this paper include regular and adversarial training and testing adversarial attack success analysis \cite{su2018robustness}. For the classification experiments, we introduce four scenarios: testing against a model trained only on regular data, using adversarial test examples for a model, and checking performance with adversarial training using either regular or adversarial test examples.

\begingroup
\setlength{\tabcolsep}{8pt} % Default value: 6pt
\renewcommand{\arraystretch}{1} % Default value: 1
\begin{table*}[!htb]
\caption{Adversarial algorithms used for model defense and attack experiments.}
\begin{center}
\begin{tabular}{l|l}
\hline
\textbf{Algorithm} & \textbf{Formulas} \\ \hline 
FGSM \cite{goodfellow2015fgsm}      &     
\begin{math}
x^{\prime}=x+\epsilon \cdot \operatorname{sgn}\left(\nabla_x \ell(f(x), y)\right)

\end{math}     
\\ \hline

RFGSM \cite{tramer2017rfgsm}    &      
\begin{math}
\begin{aligned}
& x_0^{\prime}=x+\alpha \cdot \operatorname{sgn}\left(\mathcal{N}\left(\mathbf{0}^n, \mathbf{I}^n\right)\right) \\
& x_{t+1}^{\prime}=\operatorname{clip}_{(x, \epsilon)}\left\{x_t^{\prime}+(\epsilon-\alpha) \cdot \operatorname{sgn}\left(\nabla_{x_t^{\prime}} \ell\left(f\left(x_t^{\prime}\right), y\right)\right)\right\}
\end{aligned}
\end{math}
\\ \hline

FFGSM  \cite{wong2020ffgsm}   &       
\begin{math}
\begin{aligned}
& x_0^{\prime}=x+\mathcal{U}(-\epsilon, \epsilon) \\
& x^{\prime}=\Pi_{\mathcal{B}(x, \epsilon)}\left\{x_0^{\prime}+\alpha \cdot \operatorname{sgn}\left(\nabla_{x_0^{\prime}} \ell\left(f\left(x_0^{\prime}\right), y\right)\right)\right\}
\end{aligned}
\end{math}
\\ \hline

MIFGSM  \cite{dong2017mifgsm}  &      
\begin{math}
\begin{aligned}
& g_{t+1}=\mu \cdot g_t+\frac{\nabla_{x_t^{\prime}} \ell\left(f\left(x_t^{\prime}\right), y\right)}{\left\|\nabla_{x_t^{\prime}} \ell\left(f\left(x_t^{\prime}\right), y\right)\right\|_1} \\
& x_{t+1}^{\prime}=\Pi_{\mathcal{B}(x, \epsilon)}\left\{x_t^{\prime}+\alpha \cdot \operatorname{sgn}\left(g_{t+1}\right)\right\}
\end{aligned}
\end{math}
\\ \hline

BIM  \cite{kurakin2016bim}     &        
\begin{math}
x_{t+1}^{\prime}=\operatorname{clip}_{(x, \epsilon)}\left\{x_t^{\prime}+\alpha \cdot \operatorname{sgn}\left(\nabla_{x_t^{\prime}} \ell\left(f\left(x_t^{\prime}\right), y\right)\right)\right\}
\end{math}
\\ \hline

PGD   \cite{madry2017pgd}    &       
\begin{math}
\begin{aligned}
& x_0^{\prime}=x+\mathcal{U}(-\epsilon, \epsilon) \\
& x_{t+1}^{\prime}=\Pi_{\mathcal{B}(x, \epsilon)}\left\{x_t^{\prime}+\alpha \cdot \operatorname{sgn}\left(\nabla_{x_t^{\prime}} \ell\left(f\left(x_t^{\prime}\right), y\right)\right)\right\}
\end{aligned}
\end{math}
\\ \hline

TPGD  \cite{zhang2019tpgd}    &     
\begin{math}
\begin{aligned}
& x_0^{\prime}=x+0.001 \cdot \mathcal{N}\left(\mathbf{0}^n, \mathbf{I}^n\right) \\
& x_{t+1}^{\prime}=\Pi_{\mathcal{B}(x, \epsilon)}\left\{x_t^{\prime}+\alpha \cdot \operatorname{sgn}\left(\nabla_{x_t^{\prime}} \ell_{K L}\left(f_\theta(x), f_\theta\left(x_t^{\prime}\right)\right)\right)\right\}
\end{aligned}
\end{math}
\\ \hline

EOTPGD  \cite{zimmermann2019eotpgd}  &     
\begin{math}
x_{t+1}^{\prime}=\Pi_{\mathcal{B}(x, \epsilon)}\left\{x_t^{\prime}+\alpha \cdot \operatorname{sgn}\left(\mathbb{E}\left[\nabla_{x_t^{\prime}} \ell\left(f\left(x_t^{\prime}\right), y\right)\right]\right)\right\}
\end{math}
\\ \hline
\end{tabular}
\label{tab:1-adversarial}
\end{center}
\end{table*}
\endgroup

\subsection{Model Explainability}

Introduced by Simonyan et al. \cite{simonyan2014deepic}, the saliency explanation mechanism is an attribution method based on gradients. In this method, each gradient denotes the impact of a change in each input dimension on predictions within a small neighborhood around the input. As a result, the technique calculates a class-level saliency map for an image based on the gradient of an output neuron concerning the input. This map highlights the areas in the input image, representing the focus regions for the specified class.

Guided backpropagation \cite{springenberg2014conv}, alternatively referred to as guided saliency, employs a novel deconvolution variant  \cite{zeiler2013visconv} to visualize features learned by various DNN architectures. When applied to small images, using max-pooling in CNNs becomes questionable. Replacing the max-pooling layers with a convolutional layer featuring an increased stride demonstrates no loss of performance on several image processing tasks, underscoring the versatility of guided backpropagation in challenging the conventional use of max-pooling within CNNs for specific image sizes.

Gradient-weighted class activation
mapping (GradCAM), introduced by Selvaraju et al. \cite{selvaraju2017gradcam}, addresses a fundamental limitation of class activation
mapping (CAM) \cite{zhou2016cam} by offering visual explanations for any CNN, irrespective of its architecture. This gradient-based method utilizes class-specific gradient information from the convolutional layer of a CNN to generate a localization map,  highlighting the visual attention regions in an image for classification and enhancing the transparency of CNN-based models.
Despite its advancements over CAM, GradCAM has certain limitations, such as not being able to localize multiple occurrences of an entity in an image effectively \cite{chattopadhay2018gradcampp}.
%Moreover, GradCAM may also experience a signal loss due to the iterative upsampling-downsampling operations.
% Furthermore, because of the repetitive upsampling and downsampling operations, GradCAM can suffer from a signal loss \cite{chattopadhay2018gradcampp}.

We interpret the visual performance of the models trained in different setups with visualization methods such as t-SNE \cite{hinton2002tsne} for highlighting the classifier capabilities through two-dimensional clustering, GradCAM, Guided GradCAM \cite{selvaraju2017gradcam}, and Guided GradCAM HiRes \cite{draelos2021use}. Thus, we also determine the saliency maps linked to the output of the feature extractor.

\subsection{Knowledge Distillation} 
A practical method for compressing models is knowledge distillation, which facilitates information transfer from a large teacher network to a small student network. Initially introduced by Bucila et al. \cite{bucila} and later generalized by Hinton et al. \cite{hinton2015distil}, this method has gained popularity across multiple machine learning applications. Unlike conventional training, knowledge distillation involves learning the student network to emulate the outputs of the teacher model, typically represented as probability distributions over classes. By minimizing the similarity loss between the student's predictions and the teacher's probabilities, the student network can assimilate the knowledge from the teacher, resulting in superior performance compared to training from scratch without a teacher \cite{avram2022distilling}.

% In our experiments, the student model employed in the knowledge distillation process is denoted as \textit{simple}.
The architecture of our student consists of multiple stacked convolutional layers with interleaved pooling layers and ReLU activation functions. The classification head of the student model comprises a fully-connected layer followed by a dropout layer with a dropout rate of 50\%. The utilization of dropout aids in enhancing the generalization capabilities of the model. The Adam optimizer is adopted for conducting the knowledge distillation experiments. Three teacher models were selected for our goals: ResNet50, DenseNet121 \cite{huang2017densely}, and Xception \cite{chollet2017xception}. Additionally, an ensemble approach is employed to leverage the knowledge acquired by these pre-trained teachers on the ImageNet benchmark \cite{deng2009imagenet}. Before beginning the effective distillation process, each teacher model is fine-tuned on the leaf disease classification task.

For the distillation process, the hyperparameters are crucial in controlling the knowledge transfer from the teacher models to the student model. Their values are chosen as follows: the alpha parameter is set to 0.1, and the temperature parameter is set to 3.

\subsection{Performance Metrics} 
We use accuracy and macro-F1 as performance metrics for adversarial training and attack experiments. 
Knowledge distillation experiments are evaluated using accuracy and macro-F1  for classification, the number of model parameters for model size comparison, and floating-point operations per second (FLOPs) for model efficiency evaluation.

\subsection{Research Questions} 
\label{RQ1} [RQ1] Is robustness the cost of accuracy?

\label{RQ2}\noindent [RQ2] How is the classification affected when we use adversarial examples?

\label{RQ3}\noindent  [RQ3] Does the leaf shape matter?

\label{RQ4}\noindent  [RQ4] How do adversarial attacks change focus?

\label{RQ5}\noindent  [RQ5] Is knowledge distillation effective for leaf disease classification?

\vspace{-5pt}
\section{\uppercase{Results}}
% The outcomes of our experiments are analyzed in the first two subsections in terms of classification impact on adversarial training and attack testing, followed by two subsections exploring the testing explainability of a model using multiple class activation map algorithms. We end with a knowledge distillation part, investigating the performance of the distilled models and comparing them with larger models in terms of accuracy, size, and efficiency.

\subsection{Results for RQ1}
Table~\ref{tab:2-1-accuracy} shows outstanding performance for the regular scenario, without adversarial training and adversarial attack, with 97.78\% accuracy. In contrast, when we attack the network, the actual label of adversarial examples is hardly identified, thus a lower accuracy ranging from 0.78\% to 18.58\%. 

When we train the network with adversarial examples, we observe the trade-off between robustness and accuracy \cite{su2018robustness}. Table~\ref{tab:2-2-accuracy} shows that for a network trained on adversarial examples and tested on the initial data, we determine lower accuracy for all the models, ranging from 77.17\% to 94.70\%. In the attack scenario for adversarially trained models, we gain a robustness improvement for all algorithms. BIM is the best adversarial algorithm in both types of tests, with 94.70\% accuracy for initial test data and 83.46\% for adversarial test data.

\setlength{\tabcolsep}{6pt} % Default value: 6pt
\renewcommand{\arraystretch}{1.2} % Default value: 1
\begin{table}[!ht]
\begin{center}
\caption{Accuracy (Acc) and macro-F1 (F1) for our CNN model without adversarial training.}
\begin{tabular}{c|l|l|l}
\hline
\multicolumn{1}{l|}{\textbf{Attack}}  & \textbf{Algorithm} & \textbf{Acc (\%)} & \textbf{F1 (\%)}     \\ \hline 
No              & -            & \textbf{97.75}   & \textbf{97.03} \\ \hline 
Yes & FGSM      & 5.64            & 4.69          \\ 
                                   & RFGSM     & 0.78            & 1.17          \\  
                                   & FFGSM     & 6.02            & 4.99          \\  
                                   & MIFGSM    & 0.11            & 0.17          \\  
                                   & BIM       & 10.95            & 9.24          \\  
                                   & PGD       & 0.51            & 0.76          \\  
                                   & TPGD      & \textbf{18.58}   & \textbf{17.81} \\  
                                   & EOTPGD    & 0.81            & 1.20          \\ \hline 
\end{tabular}
\label{tab:2-1-accuracy}
\end{center}
\end{table}

\vspace{-3pt}
\setlength{\tabcolsep}{5pt} % Default value: 6pt
\renewcommand{\arraystretch}{1.2} % Default value: 1
\begin{table}[!ht]
\begin{center}
\caption{Accuracy (Acc) and macro-F1 (F1) for our CNN model with adversarial training.}
\begin{tabular}{c|l|l|l}
\hline
\multicolumn{1}{l|}{\textbf{Attack}}  & \textbf{Algorithm} & \textbf{Acc (\%)} & \textbf{F1 (\%)}     \\ \hline 
No & FGSM      & 79.98            & 70.06          \\  
                                   & RFGSM     & 83.99            & 75.52          \\  
                                   & FFGSM     & 83.46            & 74.89          \\  
                                   & MIFGSM    & 77.17            & 66.84          \\  
                                   & BIM       & \textbf{94.70}   & \textbf{91.86} \\  
                                   & PGD       & 83.52            & 74.09          \\  
                                   & TPGD      & 83.37            & 77.36          \\  
                                   & EOTPGD    & 84.36            & 76.53          \\ \hline 
Yes     & FGSM      & 56.12            & 41.93          \\  
                                   & RFGSM     & 60.79            & 47.53          \\  
                                   & FFGSM     & 61.06            & 47.32          \\  
                                   & MIFGSM    & 51.29            & 36.92          \\  
                                   & BIM       & \textbf{83.46}   & \textbf{76.58} \\  
                                   & PGD       & 51.31            & 39.79          \\  
                                   & TPGD      & 64.76            & 57.78          \\  
                                   & EOTPGD    & 60.99            & 47.93          \\ \hline
\end{tabular}
\label{tab:2-2-accuracy}
\end{center}
\end{table}

\vspace{-12pt}
\subsection{Results for RQ2}
\begin{figure*}[!ht]
  \centering
  \includegraphics[width = 450pt, height = 294pt]{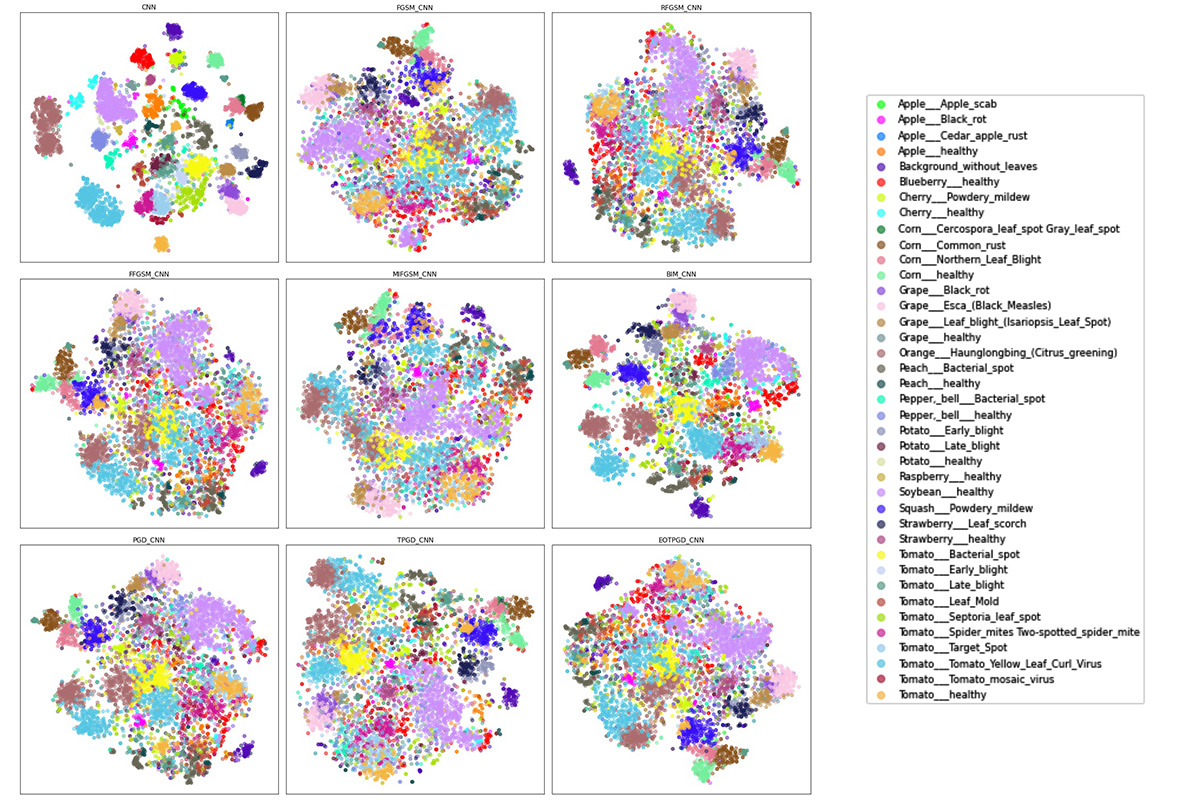}
  \caption{Class t-SNE distribution for the models trained with adversarial training. The first subfigure, labeled "CNN", depicts the classifications of the model trained without adversarial examples, while subfigures 2-9 show the classification clustering for the models using the FGSM, RFGSM, FFGSM, MIFGSM, BIM, PGD, TPGD, and EOTPGD algorithms to create more training data.}
  \label{fig:2-tsne}
\end{figure*}

Figure~\ref{fig:2-tsne} depicts the classification performance for the models as a 2D projection. We see a clear separation for the model trained and tested on initial data. Moreover, in contrast with the other adversarial models, the t-SNE plot also highlights improved performance for the BIM algorithm. Classes Orange with Huanglongbing (brown), Healthy soybean (lavender), and Tomato with yellow leaf curl virus (blue) have the most considerable volume in this dataset. Despite the correct classification, we can observe that their distribution is highly affected by adversarial training, even in the case of BIM, where the Tomato with yellow leaf curl virus is split into two separate clusters. We can still determine a slight separation for this class in regular training and testing scenarios. Thus, we conclude that adversarial training significantly affects class distributions.

\vspace{-3pt}

\subsection{Results for RQ3}

\begin{figure*}[!htb]
  \centering
  \includegraphics[width=1\textwidth]{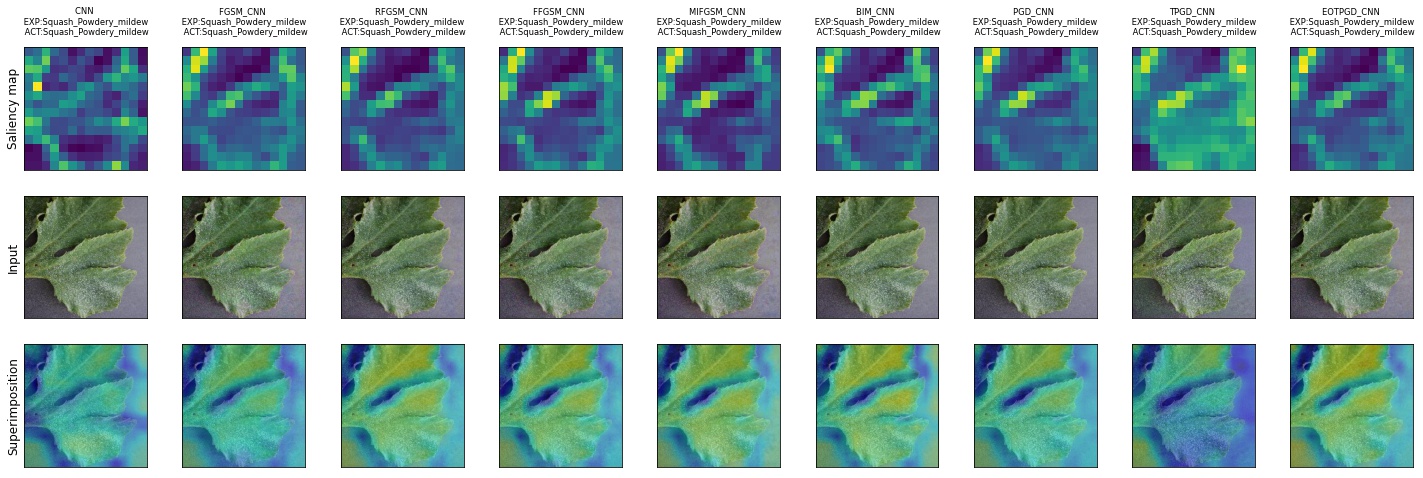}
  \caption{Squash GradCAM saliency maps without adversarial attack.}
  \label{fig:3-1-grad}
\end{figure*}

\begin{figure*}[!ht]
  \centering
  \includegraphics[width=1\textwidth]{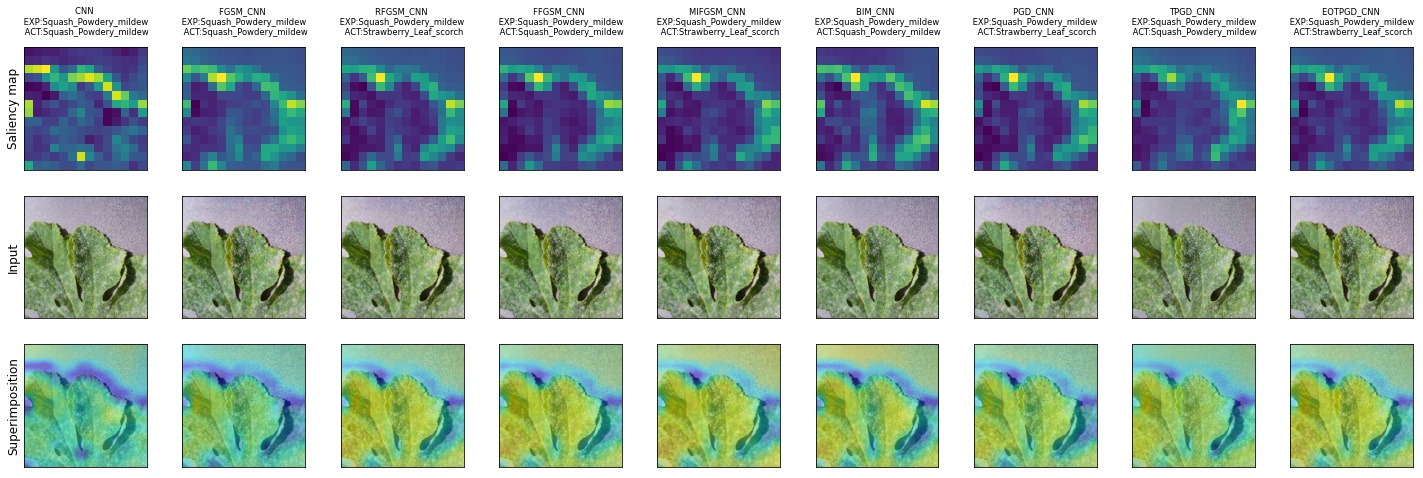}
  \caption{Squash GradCAM saliency maps with adversarial attack.}
  \label{fig:3-2-grad}
\end{figure*}

\begin{figure*}[!ht]
  \centering
  \includegraphics[width=1\textwidth]{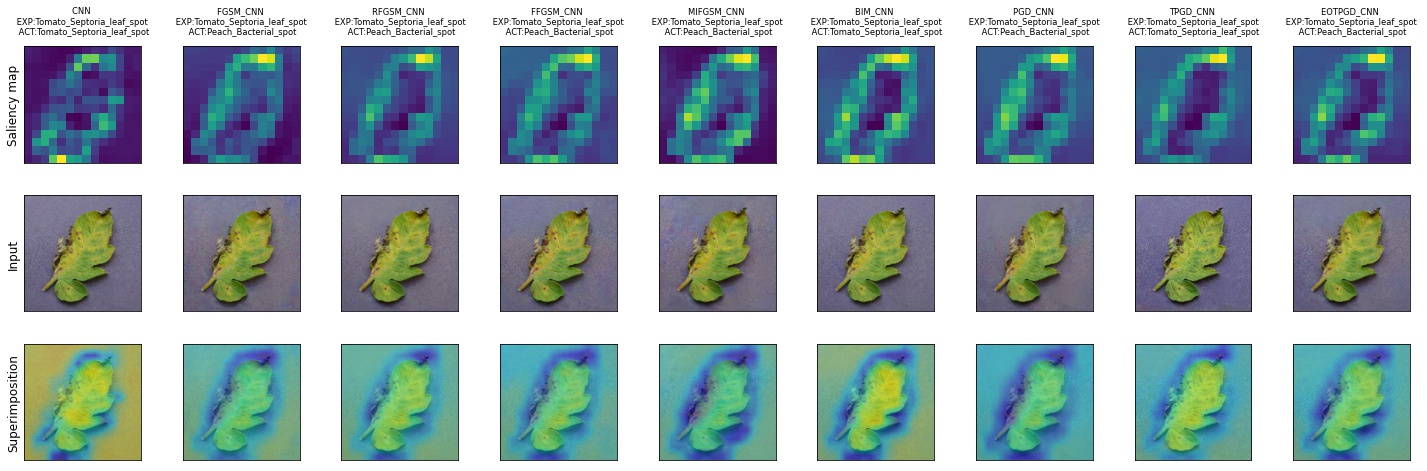}
  \caption{Tomato GradCAM saliency maps with adversarial attack.}
  \label{fig:3-3-grad}
\end{figure*}

\begin{figure*}[!ht]
\centering
\begin{subfigure}{0.48\textwidth}
    \centering
    \includegraphics[width=\textwidth]{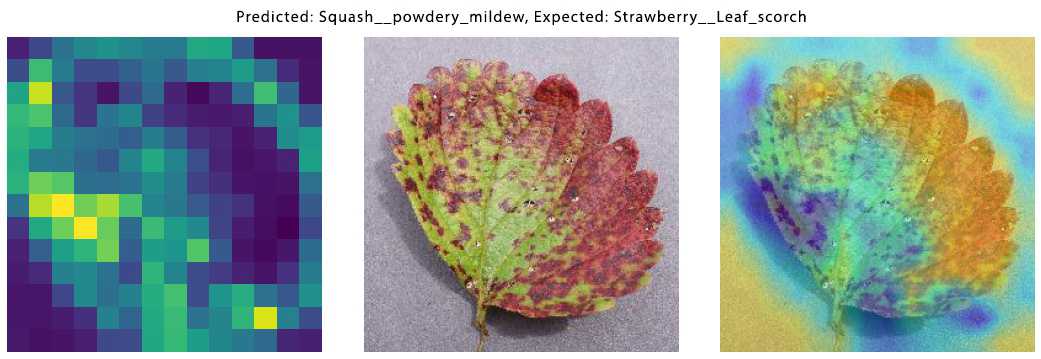}
    \caption{Strawberry classification GradCAM.}
    \label{fig:3-3-grad-justif-1}
\end{subfigure}
\hfill
\begin{subfigure}{0.48\textwidth}
    \centering
    \includegraphics[width=\textwidth]{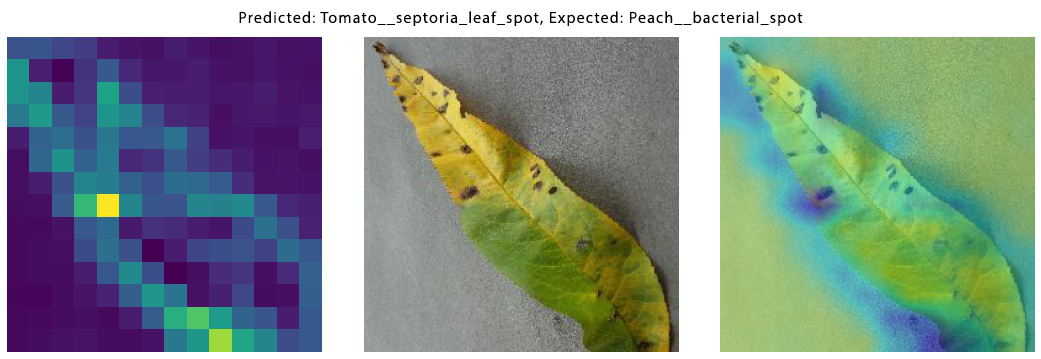}
    \caption{Peach classification GradCAM.}
    \label{fig:3-3-grad-justif-2}
\end{subfigure}
\caption{Misclassification GradCAM for different leaf diseases.}
\label{fig:3-3-grad-justif}
\end{figure*}

Figures ~\ref{fig:3-1-grad}, \ref{fig:3-2-grad}, \ref{fig:3-3-grad}, and \ref{fig:3-3-grad-justif} support an analysis focused on the classification of one image in all adversarial scenarios, providing a view on the saliency map, the test image either regular or adversarial, as well as the superimposition of them for better visualization. An initial observation is that maps for adversarial examples often describe the shape of the leaves. However, in addition to highlighting the shape and potential focus points, they feature additional heat in the regions outside the leaves due to the introduced noise. Figure~\ref{fig:3-2-grad} depicts an example where we get four misclassifications out of nine despite the similarities in shapes.

Figure~\ref{fig:3-3-grad-justif-1} attempts to explain the Strawberry with leaf scorch classification to the expected Squash with powdery mildew. The shapes of the saliency map match in significant proportion; therefore, the round features might have determined the confusion. This impact is even more evident in Figure~\ref{fig:3-3-grad}, with only three of nine correctly classified examples. A Peach with bacterial spot example is shown in Figure~\ref{fig:3-3-grad-justif-2}, featuring a similar shape and orientation to the Tomato with septoria leaf spot. We conclude that the shape and orientation of the leaves in the dataset can be prone to misclassifications when they are subject to adversarial attacks.

\subsection{Results for RQ4}

\begin{figure}[!ht]
  \centering
  \includegraphics[width=0.48\textwidth]{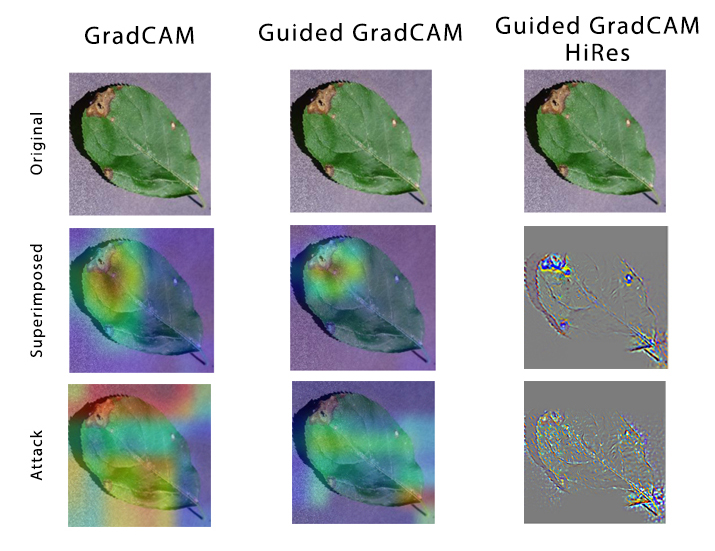}
  \caption{Model's attentional focus after the FGSM attack.}
  \label{fig:gradcam}
\end{figure}

The analysis of the model's cognitive processes during classification encompasses a comprehensive exploration of three distinct visualization techniques (i.e., GradCAM, Guided GradCAM, and Guided GradCAM HiRes), each designed to shed light on the intricate mechanisms governing its decision-making. Figure \ref{fig:gradcam} presents these visualization methods aimed at elucidating the underlying reasoning employed by the model when faced with a given instance for classification.

In particular, the second row of visualizations offers valuable insights into the model's attentional patterns when confronted with an un-attacked image. This investigation unveils crucial information regarding the areas of interest within the input image that the model prioritizes when making its predictions.

Furthermore, the last row of visualizations is a pivotal component of this analysis, seeking to reveal the precise focal points that capture the model's attention during the classification process. In this example, the FGSM attack demonstrates its efficiency by skillfully diverting the model's focus away from significant points within the leaf where the disease is present.

The ability of the FGSM attack to redirect the model's focus towards less critical regions of the image highlights potential vulnerabilities in the model's robustness and susceptibility to adversarial perturbations. Such insights are essential contributions to model explainability and adversarial machine learning, facilitating the development of more robust and dependable AI systems for critical applications.

\vspace{-0pt}

\subsection{Results for RQ5}

% The distillation process effectively enhances the performance of the student model and its ability to acquire knowledge from the teacher models. Maintaining a low number of parameters is demonstrated through the following analysis.

\textbf{Before Knowledge Distillation.} 
Table \ref{tab:before_distillation} presents the performance of the models before the knowledge distillation process. The student model achieved an accuracy of 85.38\% and a macro-F1 score of 79.17\%. In comparison, the teacher models, namely ResNet50, DenseNet121, Xception, and Ensemble, displayed significantly higher accuracies and macro-F1 scores, ranging from 97.16\% to 99.04\%.

\noindent \textbf{After Knowledge Distillation.}
Table \ref{tab:after_distillation} depicts the results after the knowledge distillation process. The student model demonstrates accuracy and macro-F1 score improvements when distilled with each teacher model. When distilled with ResNet50, the student model achieved an accuracy of 89.43\% (+4.05) and a macro-F1 score of 85.09\% (+5.92). Similarly, distillation with DenseNet121 and Xception resulted in accuracies of 90.11\% (+4.73) and 88.88\% (+3.50), as well as macro-F1 scores of 85.23\% (+6.06) and 83.42\% (+4.25). Furthermore, distillation with the Ensemble model led to an accuracy of 89.72\% (+4.34) and a macro-F1 score of 84.95\% (+5.78).

\noindent \textbf{Number of Parameters.}
Table \ref{tab:number_parameters} outlines the number of parameters for each model, both the student and teacher models. The student model contains only 3.7M parameters. In comparison, the teacher models have significantly more parameters, ranging from 7M for DenseNet121 to 51.6M for the Ensemble (comprising the parameters from ResNet50, DenseNet121, and Xception). 
Despite the knowledge transfer from the teacher models during distillation, the student model maintains a substantially lower number of parameters than the teacher models. Surprisingly, the Ensemble does not obtain the best enhancement despite the notable improvements achieved through the knowledge distillation process with various teacher models.

\noindent \textbf{Model Computational Efficiency.} \quad
The computational efficiency of the models, as indicated by the FLOPs for each one, is shown in Table  \ref{tab:flops}. The student model achieves 0.4B FLOPs. In contrast, the teacher models perform significantly more FLOPs, ranging from 7.4B for DenseNet121 to 11.9B for Xception. Even with the knowledge transfer from the teacher models during the distillation process, the student model retains its computational efficiency by operating at fewer FLOPs than the teacher models.

The results demonstrate the effectiveness of the knowledge distillation process, where the student model exhibits improved performance while acquiring knowledge from the teacher models. Moreover, the student model achieves these performance gains while keeping its parameters and computational complexity low, making it an efficient and effective solution for the leaf disease classification.

\begin{figure}[!ht]
\centering
\begin{subfigure}{0.23\textwidth}
    \centering
    \includegraphics[width=\textwidth]{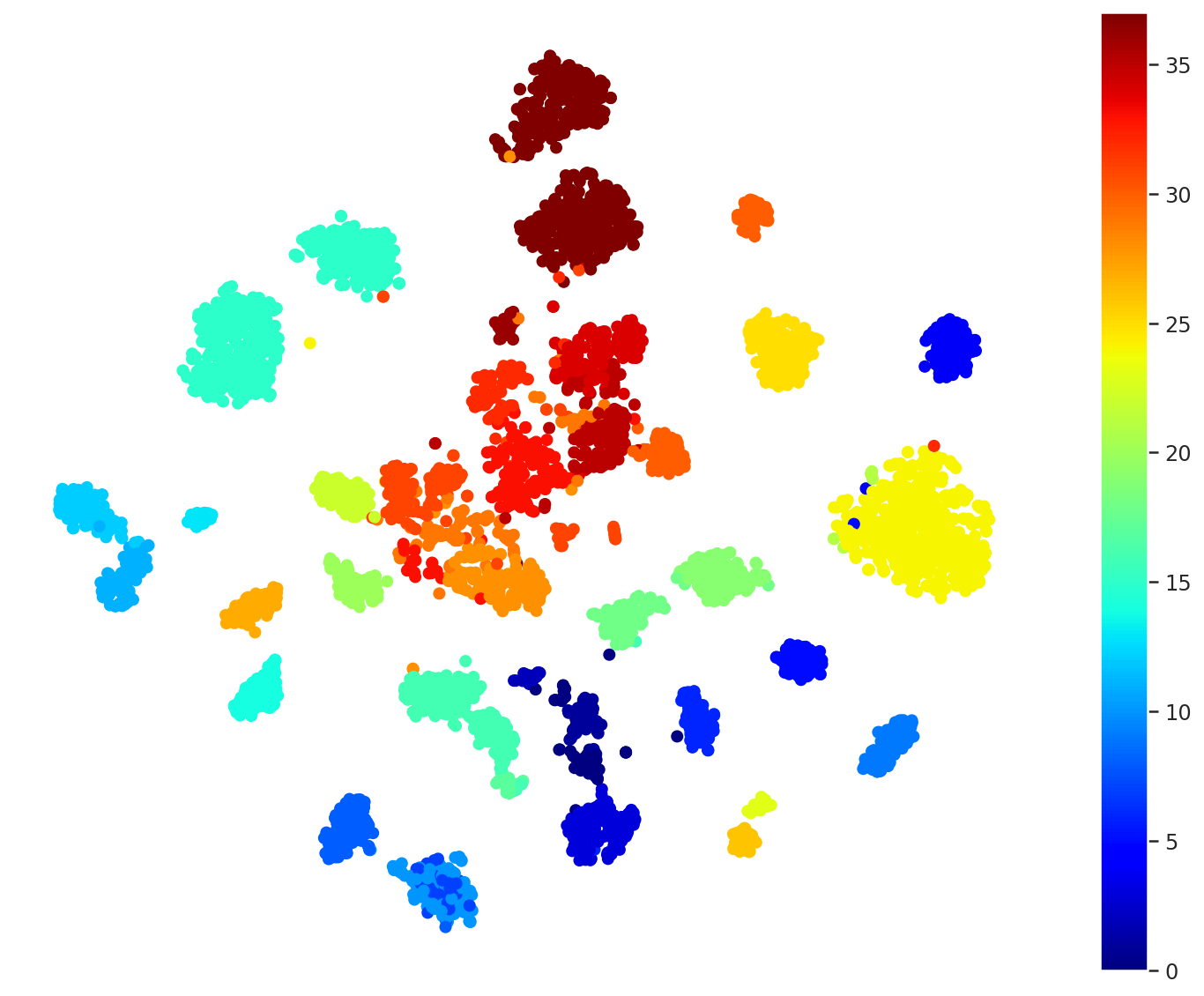}
    \caption{Teacher}
    \label{fig:teacher_tsne}
\end{subfigure}
\hfill
\begin{subfigure}{0.23\textwidth}
    \centering
    \includegraphics[width=\textwidth]{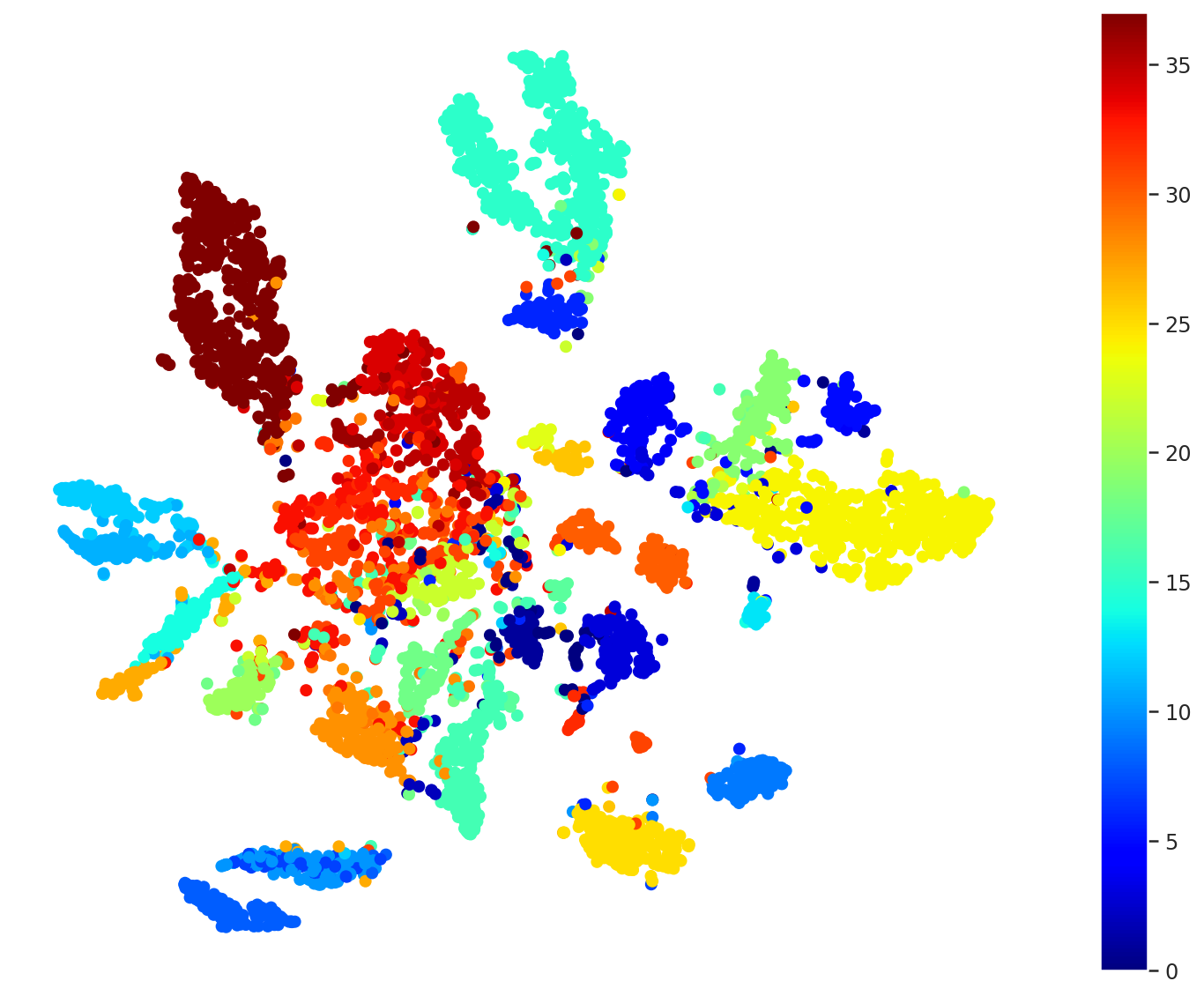}
    \caption{Student}
    \label{fig:student_tsne}
\end{subfigure}
\caption{t-SNE visualizations after knowledge distillation.}
\label{fig:teacher_student_tsne}
\end{figure}

\noindent \textbf{Student's Performance.} 
To gain insights into the student's learning progress from the teacher, we utilize t-SNE visualizations for both the teacher and the student after the knowledge distillation process. The 2D visualizations, showcased in Figure \ref{fig:teacher_student_tsne}, allow us to compare their abilities in understanding the data and clustering the leaves effectively.

The teacher demonstrates a higher capacity to comprehend the data and adeptly cluster the leaves. On the other hand, the student's performance is commendable, having learned to capture a substantial amount of knowledge from the teacher, enabling it to group the leaves accurately. However, due to the limited capacity, the student overlaps some clusters slightly.

% \begin{table}[!htb]
%   \centering
%   \caption{Performance before Knowledge Distillation}
%   \label{tab:before_distillation}
%   \begin{tabular}{l|c|c}
%     \toprule
%     Model       & Acc. (\%) & F1 (\%) \\
%     \midrule
%     Simple      & 85.38         & 79.17         \\
%     ResNet50    & \textbf{99.04}         & \textbf{98.50}         \\
%     DenseNet121 & 98.55         & 97.99         \\
%     Xception    & 97.16         & 96.34         \\
%     Ensemble    & 98.73         & 98.11         \\
%     \bottomrule
%   \end{tabular}
% \end{table}

\begin{table}[!htb]
  \centering
  \caption{Performance before knowledge distillation.}
  \label{tab:before_distillation}
  \begin{tabular}{l|c|c}
    \hline
    \textbf{Model}        & \textbf{Accuracy (\%)} & \textbf{Macro-F1 (\%)} \\
    \hline
    Student      & 85.38         & 79.17         \\
    ResNet50    & \textbf{99.04}         & \textbf{98.50}         \\
    DenseNet121 & 98.55         & 97.99         \\
    Xception    & 97.16         & 96.34         \\
    Ensemble    & 98.73         & 98.11         \\
    \hline
  \end{tabular}
\end{table}

\begin{table}[!htb]
  \centering
  \caption{Student's accuracy (Acc) and macro-F1 (F1) after knowledge distillation. }
  %For space optimization, ResNet50 is written as ``RN50", DenseNet121 as ``DN121", Xception as ``Xc", and the Ensemble of the three as "Ens". Acc is Accuracy. 
  \label{tab:after_distillation}
  \begin{tabular}{l|c|c|c|cc}
    \hline
    \textbf{Teacher} & \textbf{Acc (\%)} & \textbf{$\Delta$ Acc} & \textbf{F1 (\%)} & \textbf{$\Delta$ F1} \\
    \hline
    ResNet50    & 89.43         & +4.05          & 85.09         & +5.92           \\
    DenseNet121 & \textbf{90.11}         & \textbf{+4.73}          & \textbf{85.23}         & \textbf{+6.06}          \\
    Xception    & 88.88         & +3.50          & 83.42         & +4.25          \\
    Ensemble    & 89.72         & +4.34          & 84.95         & +5.78          \\
    \hline
  \end{tabular}
\end{table}

\begin{table}[!htb]
  \centering
  \caption{Comparison of student size with teacher models. The number of parameters is written in millions. The Ensemble model parameters are represented as the sum of the ResNet50, DenseNet121, and Xception number of parameters.}
  \label{tab:number_parameters}
  \begin{tabular}{l|c|c}
    \hline
    \textbf{Model}       & \textbf{\#Params} & \textbf{\% of Teacher Params}  \\
    \hline
    Student      & \textbf{3.71}       & -                                          \\
    ResNet50    & 23.66      & 15.68                                      \\
    DenseNet121 & 7.06       & 52.54                                      \\
    Xception    & 20.93      & 17.72                                      \\
    Ensemble    & 51.65      & 7.18                                       \\
    \hline
  \end{tabular}
\end{table}

\begin{table}[!htb]
  \centering
  \caption{Comparison of student computational efficiency with teacher models. The FLOPs values are written in gigabytes. The Ensemble model FLOPs is the maximum value of the ResNet50, DenseNet121, and Xception FLOPs.}
  \label{tab:flops}
  \begin{tabular}{l|c|c}
    \hline
    \textbf{Model}       & \textbf{FLOPs} & \textbf{\% of Teacher FLOPs} \\
    \hline
    Student      & \textbf{0.48}      & -                                         \\
    ResNet50    & 10.09     & 4.75                                      \\
    DenseNet121 & 7.45      & 6.44                                      \\
    Xception    & 11.90     & 4.03                                      \\
    Ensemble    & 11.90     & 4.03              \\
    \hline
  \end{tabular}
\end{table}

\vspace{-8pt}
\section{\uppercase{Conclusions}}
In this work, we addressed three vital pillars in machine learning: adversarial attacks, knowledge distillation, and explainability, explicitly focusing on plant leaf disease classification. Thus, we explored adversarial attacks and their impact on the robustness of CNNs. Through rigorous experimentation, we successfully employed adversarial training to enhance the resilience of the CNNs against adversarial perturbations, reinforcing their ability to perform accurate leaf disease classification.

% In machine learning interpretability, we recognized the critical importance of understanding how models arrive at their decisions, especially in the context of leaf disease classification. By delving into model explainability techniques, we gained valuable insights into the decision-making process of the CNNs, strengthening our confidence in their informed judgments and ensuring their reliability for real-world applications.

In machine learning explainability, we recognized the critical importance of understanding how models arrive at their decisions, especially in the context of leaf disease classification. Additionally, we ventured into the domain of model compression, aiming to retain the performance of the leaf disease classification models while reducing their computational complexity. Our findings demonstrated promising results, paving the way for deploying efficient and resource-friendly models, particularly crucial for applications with limited computing resources.

The following steps in our experiments include how successful adversarial attacks of an adversarial algorithm are on a model previously trained on examples generated with a different one. The quest for a better understanding of the inner workings of knowledge distillation and its interplay with other regularization techniques also opens up intriguing directions for future exploration.

\section*{\uppercase{Acknowledgements}}
This research has been funded by the National University of Science and Technology POLITEHNICA Bucharest through the PubArt program.

\bibliographystyle{apalike}
{\small
\bibliography{example}}

\begin{thebibliography}{}

\bibitem[Avram et~al., 2022]{avram2022distilling}
Avram, A.-M., Catrina, D., Cercel, D.-C., Dascalu, M., Rebedea, T., P{\u{a}}iș, V., and Tufi{\c{s}}, D. (2022).
\newblock Distilling the knowledge of romanian berts using multiple teachers.
\newblock In {\em Proceedings of the Thirteenth Language Resources and Evaluation Conference}, pages 374--384.

\bibitem[Buciluǎ et~al., 2006]{bucila}
Buciluǎ, C., Caruana, R., and Niculescu-Mizil, A. (2006).
\newblock Model compression.
\newblock In {\em Proceedings of the 12th ACM SIGKDD international conference on Knowledge discovery and data mining}, pages 535--541.

\bibitem[Chattopadhay et~al., 2018]{chattopadhay2018gradcampp}
Chattopadhay, A., Sarkar, A., Howlader, P., and Balasubramanian, V.~N. (2018).
\newblock Grad-cam++: Generalized gradient-based visual explanations for deep convolutional networks.
\newblock In {\em 2018 IEEE Winter Conference on Applications of Computer Vision}, pages 839--847.

\bibitem[Chollet, 2017]{chollet2017xception}
Chollet, F. (2017).
\newblock Xception: Deep learning with depthwise separable convolutions.
\newblock In {\em Proceedings of the IEEE conference on computer vision and pattern recognition}, pages 1251--1258.

\bibitem[Christian~Meske and Gersch, 2022]{meske2022xai}
Christian~Meske, Enrico~Bunde, J.~S. and Gersch, M. (2022).
\newblock Explainable artificial intelligence: Objectives, stakeholders, and future research opportunities.
\newblock {\em Information Systems Management}, 39(1):53--63.

\bibitem[Deng et~al., 2009]{deng2009imagenet}
Deng, J., Dong, W., Socher, R., Li, L.-J., Li, K., and Fei-Fei, L. (2009).
\newblock Imagenet: A large-scale hierarchical image database.
\newblock In {\em 2009 IEEE conference on computer vision and pattern recognition}, pages 248--255. Ieee.

\bibitem[Dong et~al., 2017]{dong2017mifgsm}
Dong, Y., Liao, F., Pang, T., Hu, X., and Zhu, J. (2017).
\newblock Discovering adversarial examples with momentum.
\newblock {\em CoRR}, abs/1710.06081.

\bibitem[Draelos and Carin, 2021]{draelos2021use}
Draelos, R.~L. and Carin, L. (2021).
\newblock Use hirescam instead of grad-cam for faithful explanations of convolutional neural networks.

\bibitem[Geetharamani and Pandian, 2019]{pandian2019data}
Geetharamani, G. and Pandian, A. (2019).
\newblock Identification of plant leaf diseases using a nine-layer deep convolutional neural network.
\newblock {\em Computers \& Electrical Engineering}, 76:323--338.

\bibitem[Ghofrani and Mahdian~Toroghi, 2022]{ghofrani2022knowledge}
Ghofrani, A. and Mahdian~Toroghi, R. (2022).
\newblock Knowledge distillation in plant disease recognition.
\newblock {\em Neural Computing and Applications}, 34(17):14287--14296.

\bibitem[Goodfellow et~al., 2015]{goodfellow2015fgsm}
Goodfellow, I.~J., Shlens, J., and Szegedy, C. (2015).
\newblock Explaining and harnessing adversarial examples.
\newblock In Bengio, Y. and LeCun, Y., editors, {\em 3rd International Conference on Learning Representations, {ICLR} 2015, San Diego, CA, USA, May 7-9, 2015, Conference Track Proceedings}.

\bibitem[He et~al., 2016]{he2016residual}
He, K., Zhang, X., Ren, S., and Sun, J. (2016).
\newblock {Deep Residual Learning for Image Recognition}.
\newblock In {\em Proceedings of 2016 IEEE Conference on Computer Vision and Pattern Recognition}, CVPR '16, pages 770--778. IEEE.

\bibitem[Hinton and Roweis, 2002]{hinton2002tsne}
Hinton, G.~E. and Roweis, S. (2002).
\newblock Stochastic neighbor embedding.
\newblock In Becker, S., Thrun, S., and Obermayer, K., editors, {\em Advances in Neural Information Processing Systems}, volume~15. MIT Press.

\bibitem[Hinton et~al., 2015]{hinton2015distil}
Hinton, G.~E., Vinyals, O., and Dean, J. (2015).
\newblock Distilling the knowledge in a neural network.
\newblock {\em CoRR}, abs/1503.02531.

\bibitem[Huang et~al., 2017]{huang2017densely}
Huang, G., Liu, Z., Van Der~Maaten, L., and Weinberger, K.~Q. (2017).
\newblock Densely connected convolutional networks.
\newblock In {\em Proceedings of the IEEE conference on computer vision and pattern recognition}, pages 4700--4708.

\bibitem[Huang et~al., 2023]{huangkd}
Huang, Q., Wu, X., Wang, Q., Dong, X., Qin, Y., Wu, X., Gao, Y., and Hao, G. (2023).
\newblock Knowledge distillation facilitates the lightweight and efficient plant diseases detection model.
\newblock {\em Plant Phenomics}, 5:0062.

\bibitem[Khan et~al., 2022]{khan2022deep}
Khan, A., Vibhute, A.~D., Mali, S., and Patil, C. (2022).
\newblock A systematic review on hyperspectral imaging technology with a machine and deep learning methodology for agricultural applications.
\newblock {\em Ecological Informatics}, 69:101678.

\bibitem[Kim, 2014]{kim2014convolutional}
Kim, Y. (2014).
\newblock Convolutional neural networks for sentence classification.
\newblock In {\em Proceedings of the 2014 Conference on Empirical Methods in Natural Language Processing (EMNLP)}. Association for Computational Linguistics.

\bibitem[Kingma and Ba, 2014]{adam2015diederik}
Kingma, D.~P. and Ba, J. (2014).
\newblock Adam: A method for stochastic optimization.
\newblock {\em arXiv preprint arXiv:1412.6980}.

\bibitem[Kurakin et~al., 2017]{kurakin2016bim}
Kurakin, A., Goodfellow, I.~J., and Bengio, S. (2017).
\newblock Adversarial examples in the physical world.
\newblock In {\em 5th International Conference on Learning Representations, {ICLR} 2017, Toulon, France, April 24-26, 2017, Workshop Track Proceedings}. OpenReview.net.

\bibitem[Linardatos et~al., 2021]{linardatos2021xai}
Linardatos, P., Papastefanopoulos, V., and Kotsiantis, S. (2021).
\newblock Explainable ai: A review of machine learning interpretability methods.
\newblock {\em Entropy}, 23(1).

\bibitem[Liu et~al., 2018]{zimmermann2019eotpgd}
Liu, X., Li, Y., Wu, C., and Hsieh, C.-J. (2018).
\newblock Adv-bnn: Improved adversarial defense through robust bayesian neural network.
\newblock In {\em International Conference on Learning Representations}.

\bibitem[Lungu-Stan et~al., 2023]{lungu2023skindistilvit}
Lungu-Stan, V.-C., Cercel, D.-C., and Pop, F. (2023).
\newblock Skindistilvit: Lightweight vision transformer for skin lesion classification.
\newblock In {\em International Conference on Artificial Neural Networks}, pages 268--280. Springer.

\bibitem[Madry et~al., 2018]{madry2017pgd}
Madry, A., Makelov, A., Schmidt, L., Tsipras, D., and Vladu, A. (2018).
\newblock Towards deep learning models resistant to adversarial attacks.
\newblock In {\em International Conference on Learning Representations}.

\bibitem[Miyato et~al., 2016]{miyato2016adversarial}
Miyato, T., Dai, A.~M., and Goodfellow, I. (2016).
\newblock Adversarial training methods for semi-supervised text classification.
\newblock {\em arXiv preprint arXiv:1605.07725}.

\bibitem[Musa et~al., 2022]{musa2022knowledgedistill}
Musa, A., Hassan, M., Hamada, M., and Aliyu, F. (2022).
\newblock Low-power deep learning model for plant disease detection for smart-hydroponics using knowledge distillation techniques.
\newblock {\em Journal of Low Power Electronics and Applications}, 12(2).

\bibitem[N{\u{a}}st{\u{a}}sescu and Cercel, 2022]{nuastuasescu2022conditional}
N{\u{a}}st{\u{a}}sescu, G.-S. and Cercel, D.-C. (2022).
\newblock Conditional wasserstein gan for energy load forecasting in large buildings.
\newblock In {\em 2022 International Joint Conference on Neural Networks (IJCNN)}, pages 1--8. IEEE.

\bibitem[Ramirez-Moreno et~al., 2013]{ramirez2013saliency}
Ramirez-Moreno, D., Schwartz, O., and Ramirez-Villegas, J. (2013).
\newblock A saliency-based bottom-up visual attention model for dynamic scenes analysis.
\newblock {\em Biological cybernetics}, 107.

\bibitem[Selvaraju et~al., 2017]{selvaraju2017gradcam}
Selvaraju, R.~R., Cogswell, M., Das, A., Vedantam, R., Parikh, D., and Batra, D. (2017).
\newblock Grad-cam: Visual explanations from deep networks via gradient-based localization.
\newblock In {\em 2017 IEEE International Conference on Computer Vision (ICCV)}, pages 618--626.

\bibitem[Simonyan et~al., 2014]{simonyan2014deepic}
Simonyan, K., Vedaldi, A., and Zisserman, A. (2014).
\newblock Visualising image classification models and saliency maps.
\newblock {\em Deep Inside Convolutional Networks}, 2.

\bibitem[Sm{\u{a}}du et~al., 2022]{smuadu2022legal}
Sm{\u{a}}du, R.-A., Dinic{\u{a}}, I.-r., Avram, A.-M., Cercel, D.-C., Pop, F., and Cercel, M.-C. (2022).
\newblock Legal named entity recognition with multi-task domain adaptation.
\newblock In {\em Proceedings of the Natural Legal Language Processing Workshop 2022}, pages 305--321.

\bibitem[Springenberg et~al., 2015]{springenberg2014conv}
Springenberg, J.~T., Dosovitskiy, A., Brox, T., and Riedmiller, M.~A. (2015).
\newblock Striving for simplicity: The all convolutional net.
\newblock In Bengio, Y. and LeCun, Y., editors, {\em 3rd International Conference on Learning Representations, {ICLR} 2015, San Diego, CA, USA, May 7-9, 2015, Workshop Track Proceedings}.

\bibitem[Su et~al., 2018]{su2018robustness}
Su, D., Zhang, H., Chen, H., Yi, J., Chen, P.-Y., and Gao, Y. (2018).
\newblock Is robustness the cost of accuracy? -- a comprehensive study on the robustness of 18 deep image classification models.
\newblock In Ferrari, V., Hebert, M., Sminchisescu, C., and Weiss, Y., editors, {\em Computer Vision -- ECCV 2018}, pages 644--661, Cham. Springer International Publishing.

\bibitem[Szegedy et~al., 2014]{szegedy2014intriguing}
Szegedy, C., Zaremba, W., Sutskever, I., Bruna, J., Erhan, D., Goodfellow, I.~J., and Fergus, R. (2014).
\newblock Intriguing properties of neural networks.
\newblock In Bengio, Y. and LeCun, Y., editors, {\em 2nd International Conference on Learning Representations, {ICLR} 2014, Banff, AB, Canada, April 14-16, 2014, Conference Track Proceedings}.

\bibitem[Tram{\`{e}}r et~al., 2018]{tramer2017rfgsm}
Tram{\`{e}}r, F., Kurakin, A., Papernot, N., Goodfellow, I.~J., Boneh, D., and McDaniel, P.~D. (2018).
\newblock Ensemble adversarial training: Attacks and defenses.
\newblock In {\em 6th International Conference on Learning Representations, {ICLR} 2018, Vancouver, BC, Canada, April 30 - May 3, 2018, Conference Track Proceedings}. OpenReview.net.

\bibitem[Wong et~al., 2020]{wong2020ffgsm}
Wong, E., Rice, L., and Kolter, J.~Z. (2020).
\newblock Fast is better than free: Revisiting adversarial training.
\newblock In {\em 8th International Conference on Learning Representations, {ICLR} 2020, Addis Ababa, Ethiopia, April 26-30, 2020}. OpenReview.net.

\bibitem[You et~al., 2023]{haotian2023adversarial}
You, H., Lu, Y., and Tang, H. (2023).
\newblock Plant disease classification and adversarial attack using simam-efficientnet and gp-mi-fgsm.
\newblock {\em Sustainability}, 15(2).

\bibitem[Zeiler and Fergus, 2014]{zeiler2013visconv}
Zeiler, M.~D. and Fergus, R. (2014).
\newblock Visualizing and understanding convolutional networks.
\newblock In Fleet, D.~J., Pajdla, T., Schiele, B., and Tuytelaars, T., editors, {\em Computer Vision - {ECCV} 2014 - 13th European Conference, Zurich, Switzerland, September 6-12, 2014, Proceedings, Part {I}}, volume 8689 of {\em Lecture Notes in Computer Science}, pages 818--833. Springer.

\bibitem[Zhang et~al., 2019]{zhang2019tpgd}
Zhang, H., Yu, Y., Jiao, J., Xing, E.~P., Ghaoui, L.~E., and Jordan, M.~I. (2019).
\newblock Theoretically principled trade-off between robustness and accuracy.
\newblock In Chaudhuri, K. and Salakhutdinov, R., editors, {\em Proceedings of the 36th International Conference on Machine Learning, {ICML} 2019, 9-15 June 2019, Long Beach, California, {USA}}, volume~97 of {\em Proceedings of Machine Learning Research}, pages 7472--7482. {PMLR}.

\bibitem[Zhou et~al., 2015]{zhou2016cam}
Zhou, B., Khosla, A., Lapedriza, {\`A}., Oliva, A., and Torralba, A. (2015).
\newblock Learning deep features for discriminative localization.
\newblock {\em 2016 IEEE Conference on Computer Vision and Pattern Recognition (CVPR)}, pages 2921--2929.

\end{thebibliography}

\end{document}